\documentclass[lettersize,journal]{IEEEtran}

\IEEEoverridecommandlockouts                              % This command is only needed if 
\newcommand{\ignore}[1]{}
\usepackage{amsfonts}
\usepackage{bm}
\usepackage{stackengine}
\usepackage{amssymb}
\usepackage{amsmath}
\usepackage{graphicx}
\usepackage{float}
\usepackage{hyperref}

\usepackage{wrapfig}
\usepackage{booktabs}
\usepackage{mwe}
\usepackage{paralist}
\usepackage[shortlabels]{enumitem}
\usepackage[utf8]{inputenc}
\usepackage{epstopdf}
\usepackage{array}
\usepackage{tabularx}
\usepackage{stfloats}
\usepackage{algorithm}
\usepackage{algpseudocode}
\usepackage{fancyhdr}
\usepackage{xcolor}
\usepackage{subcaption}
\usepackage{cite}
\usepackage{multirow}

\newtheorem{remark}{Remark}

\newcommand{\xref}{x_{\mathtt{ref}}}

\newcommand{\pobs}{p_{\mathtt{obs}}}
\newcommand{\dsafe}{d_{\mathtt{safe}}}

\newcommand{\Pb}{\ensuremath{\mathbb{P}}}

\newcommand{\Eb}{\ensuremath{\mathbb{E}}}
\newcommand{\Rb}{\ensuremath{\mathbb{R}}}
\newcommand{\HH}{\ensuremath{\mathcal{H}}}
\newcommand{\XX}{\ensuremath{\mathcal{X}}}
\newcommand{\MM}{\ensuremath{\mathcal{M}}}
\newcommand{\FF}{\ensuremath{\mathcal{F}}}

\newcommand{\PP}{\ensuremath{\mathcal{P}}}
\newcommand{\YY}{\ensuremath{\mathcal{Y}}}
\newcommand{\GG}{\ensuremath{\mathcal{G}}}

\newcommand{\innerRKHS}[2]{\ensuremath{\langle #1, #2 \rangle}_{\HH_{\XX}}}

\allowdisplaybreaks

\title{\LARGE \bf
Distributionally Robust Safe Motion Planning with Contextual Information}
\author{Kaizer Rahaman, Simran Kumari and Ashish R. Hota% <-this % stops a space
\thanks{The authors are with the Department of Electrical Engineering, IIT Kharagpur, India. Email: utoautokai@gmail.com, simranraut221151@gmail.com, ahota@ee.iitkgp.ac.in}
}

\begin{document}

\maketitle
\thispagestyle{empty}
\pagestyle{empty}

%%%%%%%%%%%%%%%%%%%%%%%%%%%%%%%%%%%%%%%%%%%%%%%%%%%%%%%%%%%%%%%%%%%%%%%%%%%%%%%%
\begin{abstract}
We present a distributionally robust approach for collision avoidance by incorporating contextual information. Specifically, we embed the conditional distribution of future trajectory of the obstacle conditioned on the motion of the ego agent in a reproducing kernel Hilbert space (RKHS) via the conditional kernel mean embedding operator. Then, we define an ambiguity set containing all distributions whose embedding in the RKHS is within a certain distance from the empirical estimate of conditional mean embedding learnt from past data. Consequently, a distributionally robust collision avoidance constraint is formulated, and included in the receding horizon based motion planning formulation of the ego agent. Simulation results show that the proposed approach is more successful in avoiding collision compared to approaches that do not include contextual information and/or distributional robustness in their formulation in several challenging scenarios. 
\end{abstract}

%%%%%%%%%%%%%%%%%%%%%%%%%%%%%%%%%%%%%%%%%%%%%%%%%%%%%%%%%%%%%%%%%%%%%%%%%%%%%%%%
\section{Introduction}
\label{sec:introduction}

Safe motion planning in uncertain and dynamic environments is a fundamental challenge in the field of robotics and autonomous systems \cite{mohanan2018survey,brunke2022safe}. Past works have proposed a plethora of approaches, including sampling-based methods \cite{orthey2023sampling}, collision avoidance using velocity obstacles \cite{fuad2020collision}, and optimization-based techniques \cite{schouwenaars2001mixed,zhang2020optimization}. In recent years, optimal control with a receding horizon implementation, popularly known as model predictive control (MPC), have gained popularity as it enables the designer to systematically include the effects of robot dynamics, constraints on robot states and inputs as well as collision avoidance constraints while planning the motion of the robot.

While early works on MPC based motion planning focused on avoiding static obstacles \cite{dixit2019trajectory,zhang2019trajectory,zhang2020optimization}, more recent works have considered dynamic obstacles in the robust and stochastic MPC frameworks. While robust MPC tends to result in conservative solutions, stochastic MPC techniques formulate collision avoidance conditions in terms of probabilistic constraints or via suitable risk measures \cite{dixit2021risk}. However, the safety guarantees provided by the above techniques may not hold when the probability distribution of the future obstacle position is unknown and/or time-varying. 

Consequently, several recent works have proposed distributionally robust motion planning techniques where the stochastic collision avoidance constraints hold for an entire family of distributions (or ambiguity set) that are close (in terms of the Wasserstein distance) to the empirical distribution constructed from the observed data \cite{hakobyan2019risk,summers2018distributionally,navsalkar2023data,zolanvari2023wasserstein,safaoui2024distributionally}. However, not all available past data are equally relevant at a given context or scenario. In particular, when we consider multiple agents that share the same environment, and are in the vicinity of each other, the future position of the obstacle is a function of the future position of the ego agent. Therefore, in order to obtain more accurate solutions, it is essential to consider ambiguity sets that depend on the actions of the ego agent and other available contextual information. Despite its significance, there is no prior work which considers decision-dependent ambiguity sets or any other contextual information while formulating the distributionally robust safe motion planning problem. 

A few recent works have explicitly included the behavior of the ego vehicle for trajectory prediction. Specifically, \cite{song2020pip} investigated the impact of ego vehicle planning on nearby vehicles' trajectories using an LSTM-based encoder combined with a convolutional social pooling module. The authors in \cite{wang2023wsip} modeled each vehicle as a wave characterized by amplitude and phase, proposing that wave-pooling better captures dynamic states and high-order interactions through wave superposition.  More recently, \cite{peng2024epn} provided a detailed comparison of several contextual trajectory prediction techniques. However, the robustness of neural network based trajectory prediction to distribution shifts is not adequately explored in the literature. Neural network based techniques also do not provide any rigorous (e.g., finite-sample) guarantees on the probability of collision among vehicles.

In this paper, we aim to fill this research gap. First, we compute the empirical estimate of the conditional kernel mean embedding (CKME) \cite{muandet2017kernel} of the (conditional) distribution of the future obstacle position as a function of the (i.e., conditioned on the) current states and the predicted trajectory of the ego agent. Then, we formulate an optimal control problem that embeds the ego-conditioned predicted trajectory in the constraints. In particular, the CKME operator provides a closed-form expression of the future trajectory of other vehicles as a function of the states and inputs of the ego vehicle, i.e., the decision variables of the optimal control problem. In order to robustify against distribution shifts, the conditional value at risk of the collision avoidance constraint is required to hold for all distributions whose mean embeddings are within a specified maximum-mean discrepancy (MMD) distance from the empirical estimate of the CKME; a similar approach was recently examined in \cite{romao2023distributionally} in the context of optimal control. Another recent paper \cite{sharma2023hilbert} explored RKHS to select probable trajectories and adopted a sampling-based optimization approach, which is distinct from our approach. We present a tractable approximation of the above constraints following the reformulations developed in \cite{zhu2021kernel,nemmour2022maximum} such that off-the-shelf nonlinear optimization solvers, such as IPOPT, can be deployed to solve the optimal control problem.\footnote{Neural network based predictors do not provide a simple closed form expression of the predicted trajectories as a function of the input applied to the ego vehicle. Therefore, computing optimal control inputs with predicted trajectories used in the constraints is not straightforward. } We provide detailed simulation results involving an autonomous ground vehicle moving on a road with obstacle agent(s), and show that the proposed approach successfully avoids collision in several challenging scenarios.

\section{Preliminaries}

\subsubsection*{Notation}
The set of real numbers is denoted by $\Rb$, with $\Rb^n$ being the $n$-dimensional Euclidean space. We define the set $[N] := \{1,2,\ldots,N\}$. The collection $\{x_k,x_{k+1},\ldots,x_{k+N}\}$ is denoted by $x_{k:k+N}$. The weighted norm $\|x\|^2_Q := x^\intercal Q x$ for a positive semidefinite matrix $Q$. The conditional value at risk of a random variable $X$ with distribution $\Pb$ at level $\alpha$ is denoted by $\mathtt{CVaR}^{\Pb}_{\alpha}[X]$. We define $(a)_{+} := \max(a,0)$. The $j$-th element of a vector $x$ is denoted by $(x)_j$. The identity function is defined as $\mathcal{I}(x)=x$. 

\subsubsection*{Background on RKHS}
Consider two measurable spaces $(\XX,\FF_{\XX})$ and $(\YY,\FF_{\YY})$. Let the set of distributions over these spaces be $\PP(\XX)$ and $\PP(\YY)$, respectively. Let $k_{\XX}$ and $k_{\YY}$ be the positive semidefinite kernels associated with both spaces, with $\HH_{\XX}$ and $\HH_{\YY}$ be the corresponding reproducing kernel Hilbert spaces (RKHS). Let $T: \YY \to \PP(\XX)$ be a measurable mapping, i.e., for every $y \in \YY$, $T(\cdot|y)$ defines a probability distribution over $\XX$. The {\bf conditional kernel mean embedding} (CKME) of $T$ is a mapping $\Psi: \YY \to \HH_{\XX}$ such that for any $f \in \HH_{\XX}$, 
$$ \innerRKHS{\Psi(y)}{f} = \Eb_{X \sim T(\cdot|y)} [f(X)]. $$

In other words, if we learn the mapping $\Psi(y)$, we can compute conditional expectation via RKHS inner product. Suppose we are not aware of the distributions of $X$ and $Y$. Instead, we have access to i.i.d. samples $\{\hat{x}_i,\hat{y}_i\}^m_{i=1}$ drawn from the joint distribution of $X$ and $Y$. Following \cite{muandet2017kernel}, the empirical estimate of the CKME is given by
$$ \widehat{\Psi}(y)(\cdot) = \sum^m_{i=1} \beta_i(y) k_{\XX} (\hat{x}_i,\cdot), $$
where 
\begin{equation}\label{eq:empirical_ckme_def}
    \beta(y) = (K_{\YY} + m\lambda I_m)^{-1} k_{y}(y) 
\end{equation}
with $K_{\YY}$ is the gram matrix with $[K_{\YY}]_{i,j} = k_{\YY}(\hat{y}_i,\hat{y}_j)$, $k_{y}(y) \in \Rb^m$ with its $i$-th entry given by $k_{\YY}(\hat{y}_i,y)$, $\lambda$ is the regularization parameter and $I_m$ is the identity matrix. Thus, for any $f \in \HH_{\XX}$, we have $\innerRKHS{\widehat{\Psi}(y)}{f} = \sum^m_{i=1} \beta_i(y) f(\hat{x}_i)$. 

\section{Problem Formulation}
\label{sec:system_description}

In this section, we formally introduce the problem of safe motion planning in presence of dynamic obstacles, and propose the distributionally robust collision avoidance approach leveraging contextual information. We consider the kinematic bicycle model \cite{kong2015kinematic} where the point of reference for the ego vehicle is considered to be the center of the rear axle. The state of the ego vehicle is defined by four components: \( p_x \) denotes the $x$-coordinate of the vehicle, \( p_y \) denotes the $y$-coordinate of the vehicle, \( \theta \) denotes the orientation angle with respect to the $x$-axis, and \( v \) denotes the velocity of the vehicle. The inputs to the bicycle model are the acceleration of the ego vehicle $a$, and the steering angle of the front wheels $\delta$. 

Let $x:=[p_x,p_y,\theta,v]^\intercal \in \mathbb R^4$ denote the state vector and $u:=[a,\delta]^\intercal \in \mathbb R^2$ denote the input vector. The equations of motion for the ego agent can be expressed as
\begin{subequations}\label{eq:bicycle_model}
\begin{align}
    \dot{p_x} &= v \cos(\theta), \\
    \dot{p_y} &= v \sin(\theta), \\
    \dot{\theta} &= \frac{v}{L} \tan(\delta), \\
    \dot{v} &= a,
\end{align}
\end{subequations}
where $L$ is the distance between the front and rear axles of the ego agent, known as the wheelbase. We discretize the above dynamics using the forward Euler method\footnote{ We have adopted the Euler discretization scheme due to its simplicity and given that it is widely adopted in the motion planning literature \cite{lindemann2023safe, smit2022informed, nair2024predictive}.} with sampling time $T_s$, and compactly represent the discrete-time dynamics as
\begin{align}
    x(k+1)=f_d(x(k),u(k)),
\end{align}
where $k$ denotes time instant $kT_s$. We will often denote $p:=[p_x,p_y]^\intercal$ for better readability.

We formulate the problem of safe motion planning for the ego vehicle in the presence of a dynamic obstacle as a finite horizon optimal control problem which is solved in a receding horizon fashion. The ego vehicle needs to reach a target state $\xref$ while avoiding the space occupied by the obstacle during maneuver. The target state could indicate either a desired target position that the ego vehicle needs to reach or a desired velocity that the ego vehicle needs to maintain. We assume that there is another vehicle in the same environment which acts as an obstacle for the ego vehicle. Safety is achieved by maintaining a minimum (Euclidean) distance between the rear axle centers of the vehicles. 

Without loss of generality, let the current time be $k=0$, and the current state $x(0)$ and previous input $u(-1)$ be known. The safe motion planning problem is formulated as:
\begin{subequations}\label{eq:mpc_formulation}
\begin{align}
    \min_{u_{0:N-1}, x_{1:N}} & %\sum_{i=1}^N \! \|u(i)\!-\!u(i\!-\!1)\|_{R}^2 + \|x(N\!+\!1) \!-\! \xref\|_Q^2 \\
    \sum_{i=0}^{N-1} \Big[ \|x(i+1) \!-\! x_{\tt ref}\|_{Q_i}^2 + \|u(i)\|^2_{R_1} \label{eq:mpc_cost}
    \\ & \quad + \|u(i)\!-\!u(i\!-\!1)\|_{R_2}^2 \Big] \nonumber \\ 
    \text{s.t.} \quad 
    & x(i+1) = f_d(x(i), u(i)),  \\
    & u_{\min} \leq u(i) \leq u_{\max} \label{eq:mpc_input_constraint} \\
    & x_{\min} \leq x(i+1) \leq x_{\max}  \label{eq:mpc_state_constraint} \\
    & \mathcal{R}(p(i+1),\pobs(i+1)) \leq 0, \label{eq:mpc_safe_constraint} \\
    & x(0), u(-1) \text{ are given},
\end{align}
\end{subequations}
where $N$ is the prediction horizon and the constraints hold for all $i \in \{0,1,\ldots,N-1\}$. The first term in the cost function penalizes distance from the desired goal state over the prediction horizon, the second term penalizes control effort, the third term penalizes sudden change in control input. The matrices $Q_i,R_1, R_2$ are positive semi-definite matrices of suitable dimensions that act as weighting factors in the cost functions. The constraint \eqref{eq:mpc_input_constraint} and \eqref{eq:mpc_state_constraint} capture nominal constraints on the state and inputs. The constraint \eqref{eq:mpc_safe_constraint} guarantees that the ego vehicle does not collide with the obstacle; in particular, we assume 
\begin{equation}\label{eq:safety_constraint}
    \mathcal{R}(p(i),\pobs(i)) := \dsafe - \| p(i) - \pobs(i) \|^2_2,
\end{equation}
where $\dsafe$ is the safety margin and $\pobs(i)$ denotes the position of the obstacle at time $i$. Multiple obstacles can be included by imposing the constraint \eqref{eq:mpc_safe_constraint} for each obstacle. In practice, the future position of the obstacle is not known to the ego agent while solving \eqref{eq:mpc_formulation}. We now describe three different approaches to infer $\pobs(i)$ and solve \eqref{eq:mpc_formulation}. 

\subsubsection{Constant Velocity Assumption}
\label{subsection:constant_velocity}
In this setting, we assume that the the current state of the obstacle agent is known, and the obstacle moves with a constant velocity without changing its orientation over the prediction horizon $N$. Then, we predict the future position of the obstacle over the prediction horizon, and define the collision avoidance constraints with respect to these predicted positions. 

\subsubsection{Contextual Prediction using CKME}
\label{subsection:contextual_prediction}

While the constant velocity assumption enables us to easily compute the future position of the obstacle, this assumption is quite strong when the vehicles are close to each other. In particular, the future position of the obstacle is a function of the current and future positions of the ego vehicle. We now leverage the Conditional Kernel Mean Embedding (CKME) approach to contextualize the obstacle dynamics relative to the ego agent's motion to incorporate the interaction between the ego agent and the obstacle. In particular, we learn $N$ CKME operators to predict the position of the obstacle $i$ time-steps into the future for $i \in [N]$. The CKME operator for $i$ time-step prediction is denoted by $\GG_i: (\Rb^2)^{i+1} \mapsto \HH_{\XX}$ which takes as input the current position of the obstacle and ego vehicles as well as the change in position of the ego vehicle from $t\in\{0,1,\ldots,i-1\}$. The output of the operator is the position of the obstacle after $i$ time steps, obtained by computing the inner product of $\GG_i$ with an identity function $\mathcal{I}$. Formally, 
\begin{align}
    p^{\mathtt{cx}}_{\mathtt{obs}}(i,z_i) &= \innerRKHS{\GG_i(z_i)}{\mathcal{I}}, \label{eq:contextual_prediction} \\
    \text{where,}\quad & z_i = [p(0)^\intercal, (p(1)-p(0))^\intercal, \ldots, \nonumber
    \\ & \qquad \qquad (p(i-1)-p(0))^\intercal, \pobs(0)^\intercal ]^\intercal. \nonumber
\end{align}
The notation $p^{\mathtt{cx}}_{\mathtt{obs}}(i,z_i)$ makes the dependence on $z_i$ explicit, while the superscript $\mathtt{cx}$ indicates that the prediction uses contextual information via CKME. 

Let the dataset available to us be denoted by $\mathcal{D}_i := \{(\hat{z}^{(1)}_i, \widehat{p}^{(1)}_{\mathtt{obs}}(i)), \ldots, (\hat{z}^{(N_s)}_i, \widehat{p}^{(N_s)}_{\mathtt{obs}}(i))\}$ which contains $N_s$ input-output pairs. The empirical CKME obtained from the above dataset is given by
\begin{equation*}
    \GG_i(z_i)(\cdot) = \sum^{N_s}_{j=1} \beta^j_i(z_i;\mathcal{D}_i) k_{\XX} (\widehat{p}^{(j)}_{\mathtt{obs}}(i),\cdot)
\end{equation*}
where $\beta^j_i(z_i;\mathcal{D}_i)$ are the coefficients that depend on the context $z_i$ as well as the past data $\mathcal{D}_i$ according to \eqref{eq:empirical_ckme_def}. Consequently, 
\begin{align}\label{eq:contextual_prediction2}
    p^{\mathtt{cx}}_{\mathtt{obs}}(i,z_i) &= \sum^{N_s}_{j=1} \beta^j_i(z_i;\mathcal{D}_i) \widehat{p}^{(j)}_{\mathtt{obs}}(i).
\end{align}
Under this approach, the MPC problem is solved with $p^{\mathtt{cx}}_{\mathtt{obs}}(i,z_i)$ used in place of $\pobs(i)$ in \eqref{eq:safety_constraint}; \eqref{eq:contextual_prediction2} provides a closed-form expression of $\pobs(i)$ in terms of $z_i$ that are decision variables in the MPC problem. 

\subsubsection{Distributionally Robust Contextual MPC}
\label{drcmpc: distributionally robust contextual mpc}

While the predictions obtained in \eqref{eq:contextual_prediction} incorporates contextual information, these predictions may not be accurate due to a finite number of data samples used to obtain the empirical estimate of the CKME. Consequently, there could be potential violations of safety constraints. Therefore, we need to robustify the safety constraints over a family of distributions that are close to the empirical estimate of the conditional distribution approximated by the CKME. To this end, for $i \in [N]$, we define the following kernel ambiguity set which depends on contextual information as
\begin{equation}\label{eq:contextual_kernel_ambiguity}
    \MM^{\epsilon}_i(z_i) := \{ \Pb \in \PP(\XX) | \| \Psi(\Pb) - \GG_i(z_i) \|_{\HH_{\XX}} \leq \epsilon \},
\end{equation}
where $\Psi(\Pb)$ is the embedding of the distribution $\Pb$ in the RKHS $\HH_{\XX}$ \cite{muandet2017kernel}. The above set contains all distributions whose kernel mean embedding is within MMD distance $\epsilon > 0$ from $\GG_i(z_i)$. $\MM^{\epsilon}_i(z_i)$ captures distributional perturbations in a nonparametric, infinite-dimensional feature space induced by the kernel. The ambiguity set also incorporates contextual information due to its dependence on $z_i$, i.e., if the ego vehicle chooses a different sequence of control inputs, then the (center of the) ambiguity set will be different as well.

We now reformulate the safety constraint. We treat $\pobs(i)$ as a random variable $\xi_i$ whose distribution $\Pb$ belongs to the ambiguity set $\MM^{\epsilon}_i(z_i)$. Consequently, the function $\mathcal{R}(p(i),\xi_i)$ is now a random variable. We require the {\it conditional value at risk} (CVaR) of $\mathcal{R}(p(i),\xi_i)$ at a suitably chosen level $\alpha \in [0,1]$ to be upper bounded by $0$ for all distributions in $\MM^{\epsilon}_i(z_i)$. The bound on CVaR will not only guarantee that the constraint is satisfied with probability at least $1-\alpha$, it would also ensure that expected magnitude of constraint violation is small. Formally, the following constraint is added in place of \eqref{eq:mpc_safe_constraint} in \eqref{eq:mpc_formulation}:
\begin{equation}\label{eq:dro_cvar_safety}
    \sup_{\mathbb{P}\in\MM^{\epsilon}_i(z_i)} \mathtt{CVaR}^{\Pb}_{\alpha} [\dsafe - \|p(i) - \xi_i\|_2] \leq 0.
\end{equation}

Following \cite{nemmour2022maximum}, we reformulate the above constraint as
\begin{align}
    & g_o^i + \sum_{j=1}^{N_s} \beta^j_i(z_i;\mathcal{D}_i) (K_{X}^i \gamma^i)_j + \epsilon \sqrt{\gamma^{i\intercal} K_{X}^i \gamma^i} \leq t^i \alpha, \label{after_approx} \\
    & (\dsafe^2 - ||p(i) - \widehat{p}^{(j)}_{\mathtt{obs}}(i)||_2^2 + t^i)_+ \nonumber
    \\ & \qquad \qquad \qquad \qquad \qquad \leq g_o^i + (K_{X}^i \gamma^i)_j, \quad j \in [N_s]  \nonumber \\
    & \gamma^i \in \mathbb{R}^{N_s}, \quad t^i \in \mathbb{R}, \quad g_o^i \in \mathbb{R}, \nonumber
\end{align}
where $K^i_{X}$ is the gram matrix defined using points $\{\widehat{p}^{(1)}_{\mathtt{obs}}(i), \ldots, \widehat{p}^{(N_s)}_{\mathtt{obs}}(i)\}$. 

The MPC problem \eqref{eq:mpc_formulation} is solved by replacing the constraint \eqref{eq:mpc_safe_constraint} with the above set of constraints for all $i \in [N]$. 

\begin{remark} 
The ambiguity set defined in terms of the MMD distance satisfies finite-sample guarantees as stated in \cite{zhu2021kernel,nemmour2022maximum}. Specifically, let \(\mathbb{P}_0\) be the true distribution of the uncertain parameter $\xi$ and let \(\widehat{\mathbb{P}}_N\) be the empirical distribution based on \(N\) i.i.d. samples of $\xi$ drawn from $\mathbb{P}_0$. Then, the MMD distance between \(\mathbb{P}_0\) and \(\widehat{\mathbb{P}}_N\) satisfies
\begin{equation}
\label{eq:rate}
\mathtt{MMD}(\mathbb{P}_0, \widehat{\mathbb{P}}_N) \leq \sqrt{\frac{C}{N}} + \sqrt{\frac{2C \log(1/\delta)}{N}},
\end{equation}
with probability at least \(1 - \delta\) \cite{nemmour2022maximum}. In the above equation, \(C\) is a constant satisfying \(\sup_x k(x, x) \leq C < \infty\). Nevertheless, the above expression yields a conservative upper bound on the radius \(\varepsilon\). To obtain a less conservative estimate, \cite{nemmour2022maximum} presented a bootstrap-based estimation procedure which we adopt in our simulations.
\end{remark}

%Thus, for a desired $\delta$ and number of available samples $N$, we can choose $\epsilon \geq \sqrt{\frac{C}{N}} + \sqrt{\frac{2C \log(1/\delta)}{N}}$ to guarantee that the true distribution lies in the ambiguity set with probability at least $1-\delta$. 

%%%%%%%%%%%%%%%%%%%%%%%%%%%%%%%%%%%%%%%%%%%%%%%%%%%%%%%%%%%%%%%%%%%%%%%%%%%%%%%%

\section{Simulation Results}
\label{sec:results}

\begin{figure*}[htbp]
    \centering
    \includegraphics[width=0.9\textwidth]{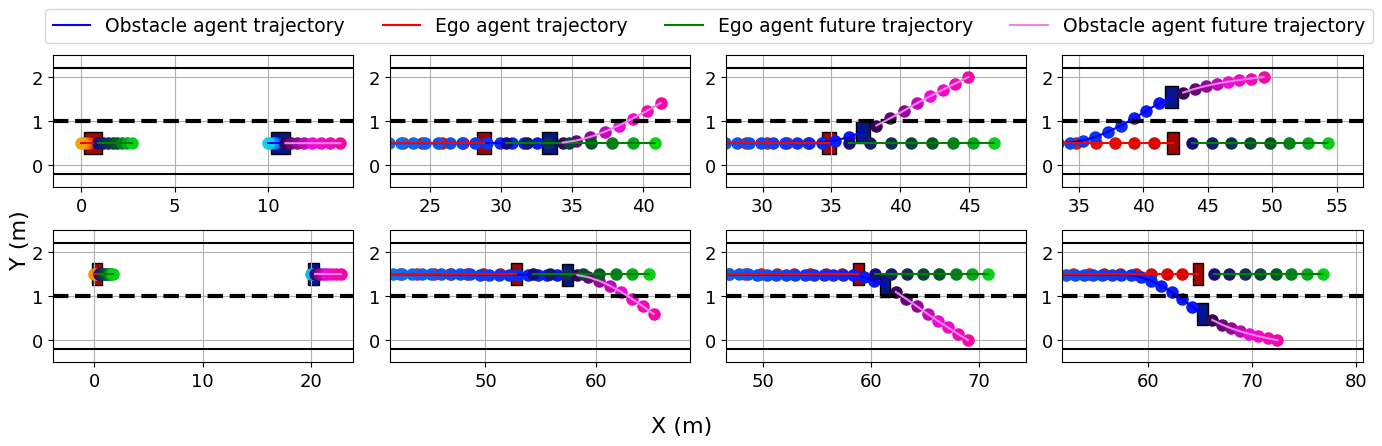}  % Adjust width as needed
    \caption{\footnotesize{Trajectories of the ego and obstacle agents during data generation. Top row captures Obstacle I and the bottom row captures Obstacle II. The obstacle agent, which moves from left to right, is shown in blue, and the non-reactive ego is shown in red.}}
    \label{fig:experience}
\end{figure*}

In this section, we demonstrate the effectiveness of the proposed approach via numerical simulations. All simulations were executed on a workstation with a 64-bit \texttt{x86\_64} architecture, equipped with an Intel(R) Core(TM) i7-10700 processor with base frequency of 2.90 GHz, and 32 GB of RAM. We utilized the Python programming environment via the Anaconda platform, employing the CasADi optimization framework~\cite{andersson2019casadi}, along with the MA57 linear solver \cite{scott2023hsl}.

\subsection{Data generation and training}
In order to generate the necessary training data set, we simulate an ego agent and an obstacle agent, with the motion of both agents governed by a model predictive controller. The ego MPC is {\it non-reactive}, i.e., it does not include the collision avoidance constraints, while the obstacle MPC is {\it reactive}, i.e., it aims to avoid collision with the ego agent under the constant velocity assumption. We consider two cases.
\begin{itemize}
    \item \textbf{Case I (Single Obstacle Agent)}: Both agents move in the same direction on the same lane. The obstacle agent is positioned ahead of the ego agent, and travels with a smaller velocity compared to the ego agent. Thus, unless the ego agent slows down or one of the agents changes its lane, there is a likelihood of collision. 
    \item \textbf{Case II (Multiple Obstacle Agents)}: Similar to Case I except that we have an additional obstacle agent positioned further ahead in the opposite lane moving with the same velocity as the other obstacle agent.
\end{itemize}

\begin{figure*}[htbp]
    \centering
    \includegraphics[width=0.8\textwidth]{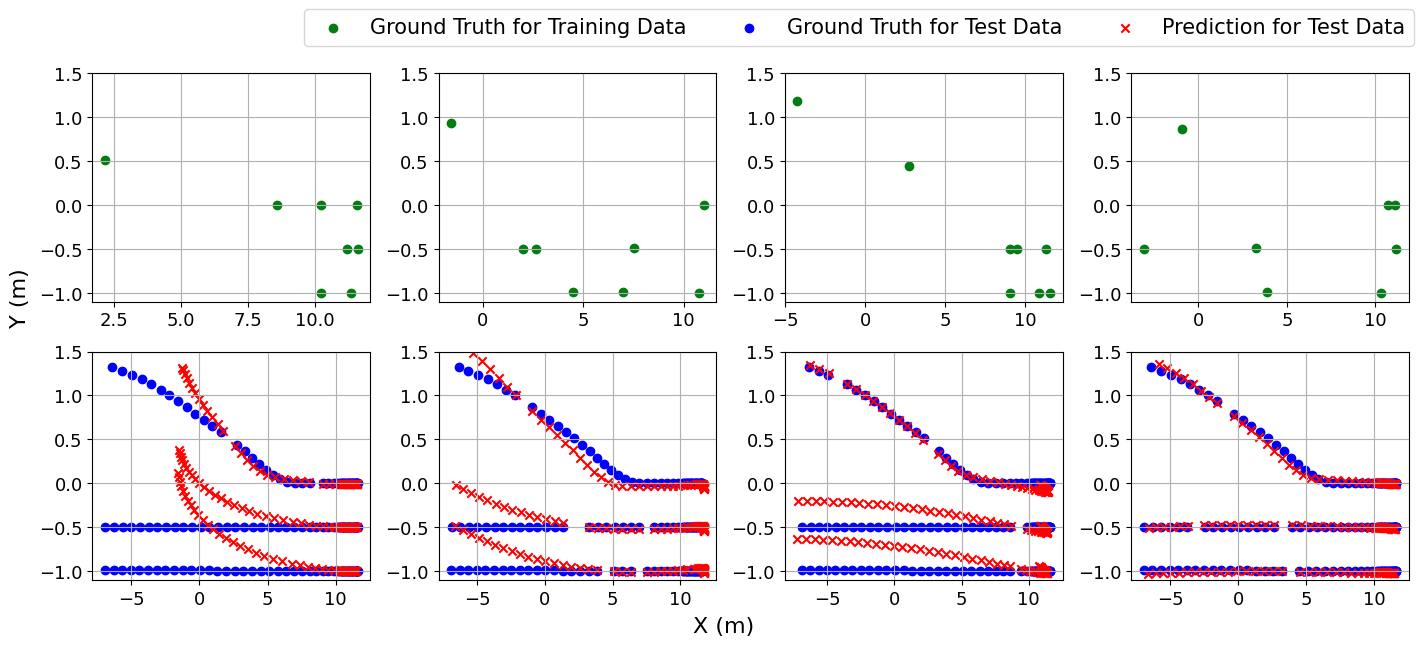}  
    % Adjust width as needed
    \caption{\footnotesize{\textbf{Variation in the fit of the test dataset under different training subsets at $i = 2$ for Obstacle I.}
    Each point represents the position of the obstacle agent relative to the ego agent. The corresponding context, $z_i$, is not shown due to its high dimensionality and lack of visual interpretability. The axes represent the $x$ and $y$ coordinates relative to the ego agent. We choose $N_s = 8$ samples for each training subset. The top row illustrates the future positions of the obstacle, $\hat{p}_{\mathtt{obs}}(i)$, for various training data selections (highlighted in green). The bottom row shows the fitted distribution over the remaining test dataset, i.e., $\mathcal{D} \setminus \{(\hat{z}_i, \hat{p}_{\mathtt{obs}}(i))\}$, corresponding to each training subset $\{(\hat{z}_i, \hat{p}_{\mathtt{obs}}(i))\}$ shown above. The ground truth obstacle position $p_{\mathtt{obs}}(i)$ is marked in blue, while the CKME prediction $p^{\mathtt{cx}}_{\mathtt{obs}}(i, z_i)$ is marked in red.}}
    \label{fig:GPR_random_state}
\end{figure*}

The data generated by the above simulations are used to construct $\mathcal{D}_i := \{(\hat{z}^{(1)}_i, \widehat{p}^{(1)}_{\mathtt{obs}}(i)), \ldots, (\hat{z}^{(N_s)}_i, \widehat{p}^{(N_s)}_{\mathtt{obs}}(i))\}$ for all $i \in [N]$ where $N=8$ is the prediction horizon. Since the obstacle agent was assumed to be reactive, its position after $i$ time-steps is a function of the position of the ego agent from time $0$ to $i-1$. We choose the radial basis function (RBF) kernels for both the input and output spaces and learn the CKME operators $\GG_i$ for all $i \in [N]$ following \eqref{eq:empirical_ckme_def}. Figure \ref{fig:experience} shows the trajectories of the ego and obstacle agents during data generation. In both scenarios, the obstacle changes its lane to avoid collision with the ego agent. The trajectory data is processed to extract the context variables and the output variables for all $i \in [N]$.

Figure \ref{fig:GPR_random_state} illustrates the fit on the entire dataset using various combinations of small, fixed-size training subsets {$\hat{z}_i$, $\hat{p}_{\mathtt{obs}}(i)$} from our generated data {$\hat{\mathbf{y}}^i$, $\hat{\mathbf{p}}_o^i$} under Scenario I. As seen in Figure \ref{fig:GPR_random_state}, a sub-optimal selection of training points leads to a less precise fit for modeling obstacle behavior. To address this, we conduct a grid search to determine the optimal set of training parameters ($\hat{z}_i$ and $\hat{p}_{\mathtt{obs}}(i)$) that interpolate effectively across the entire generated dataset. The traditional mean square error (MSE) is employed as the evaluation metric for the fit's accuracy. Notably, slight variations in the selection of training data points can significantly improve the fit’s precision, as demonstrated in the progression from left to right in Figure \ref{fig:GPR_random_state}. This approach has been used to identify the optimal training points ($\hat{z}_i$ and $\hat{p}_{\mathtt{obs}}(i)$).

We choose the \textit{Radial Basis Function} (RBF) kernel in our work,\footnote{The Matern kernel is also a popular choice in certain applications \cite{yoon2021interaction, stephens2024planning}. We compared both, and found that the predictions under the RBF kernel are more smooth compared to those under the Matern kernel.} and learn the empirical conditional mean embedding by employing the \emph{Gaussian Process} package from \emph{scikit-learn}. The length scale of the kernel defined on the output space was tuned using the median heuristic \cite{nemmour2022maximum}, and the regularization parameter is fixed at \textbf{$1 \times 10^{-4}$}. 

\subsection{Results}\label{subsection:simulation_testscenarios}

We consider three motion planning approaches for the ego vehicle. When the ego vehicle used predicted obstacle position under the constant velocity assumption, we refer to it as the nominal MPC (NMPC). When the predicted obstacle position computed using the CKME is used for collision avoidance, we refer to it as the contextual MPC (CMPC). Finally, the distributionally robust approach is referred to as the DRCMPC. We use prediction horizon $N=8$ and sampling time $T_s=0.1$ is used for Euler discretization. Table \ref{tab:mpc_comparison_scenario_I} depicts the values of the parameters of the MPC problem \eqref{eq:mpc_formulation} used in our simulations. In the cost function of the MPC problem, we use $R_1 = 0$ and $Q_i = 0$ for $i \in \{0,1,\ldots,N-2\}$. In other words, we do not penalize control effort, and only penalize rate of change of control inputs to avoids abrupt changes in acceleration or steering, which are often undesirable for passenger comfort. Similarly, we only penalize the difference between the state of the ego vehicle and the goal state at the end of the prediction horizon.

%\footnote{In many applications, penalizing control effort and deviation from the goal state along the trajectory are also important, these can be captured using the cost function defined in \eqref{eq:mpc_formulation}.} We use SI units for the states and other parameters. The confidence level associated with CVaR, $\alpha$, is set to $0.1$.

We compare the performance of each of the above three approaches against obstacles that adopt one of the following types of behavior.
\begin{itemize}
    \item {\it Non-reactive/Neutral}: The obstacle solves a MPC problem to reach a target destination and does not attempt to avoid collision with the ego agent. 
    \item {\it Cooperative}: The obstacle solves a MPC problem including collision avoidance constraints under the assumption that the ego agent moves with a constant velocity. 
    \item {\it Adversarial}: The obstacle does not include collision avoidance constraints, and changes its velocity in response to the motion of the ego agent in order to provoke collision with the ego agent. Within this case, we consider each obstacle agent of the two scenarios to follow the following behavior. For both scenarios, the change in velocity of the obstacle agent is twice that of the ego agent, i.e., 
    \begin{equation}\label{eq:scenario1_velocity}
    \Delta v_{\text{obs}}(t) = 2 \cdot \Delta v_{\text{ego}}(t).
    \end{equation}
\end{itemize}

\begin{table}[htbp]
\centering
\begin{tabular}{|c|c|c|c|c}
\hline
\textbf{Parameter} & \textbf{NMPC} & \textbf{CMPC} & \textbf{DR-CMPC} \\ \hline
$R_2$ & \textbf{diag}[0,1e-4] & \textbf{diag}[0,1e-4] & \textbf{diag}[0,1e-4] \\ \hline
%P & \textbf{diag}[0,0,0,0] & \textbf{diag}[0,0,0,0] & \textbf{diag}[0,0,0,0] \\ \hline
$Q_{N-1}$ & \textbf{diag}[0,0,0.1,0] & \textbf{diag}[0,0,0.1,0] & \textbf{diag}[0,0,0.1,0] \\ \hline
$u_{\max}$ & [4, 0.5] & [4, 0.5] & [4, 0.5] \\ \hline
$u_{\min}$ & [-4, -0.5] & [-4, -0.5] & [-4, -0.5] \\ \hline
$\dsafe$ & 0.5 & 0.5 & 0.5 \\ \hline
\end{tabular}
\caption{Values of parameters used in the MPC problem}
\label{tab:mpc_comparison_scenario_I}
\end{table}

\begin{figure*}[ht]
    \centering
    \includegraphics[width=0.9\textwidth]{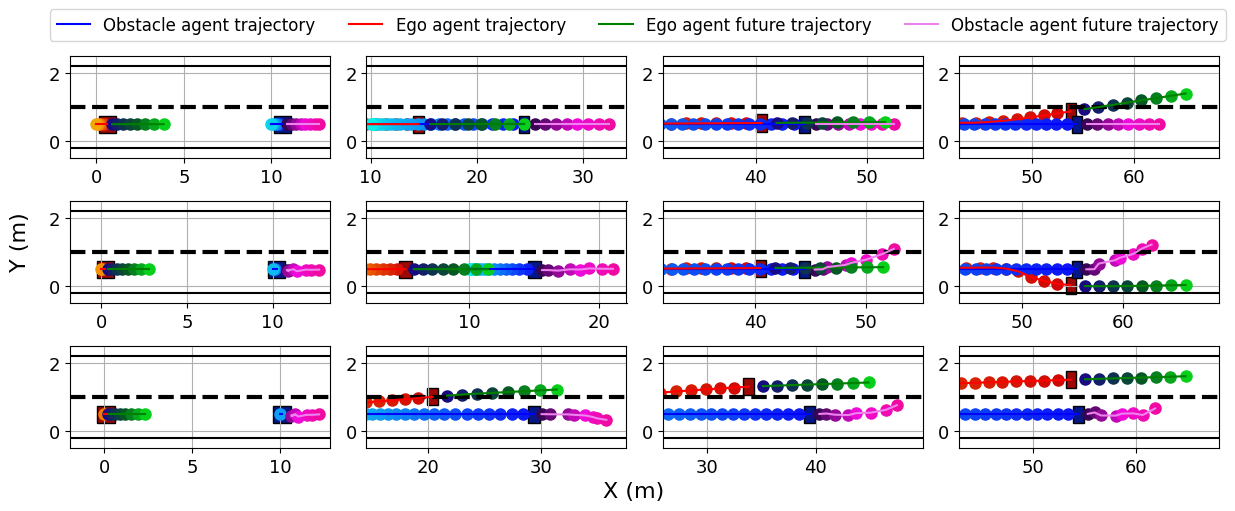}
    \caption{\footnotesize{Actual and predicted trajectories of the ego (red) and obstacle (blue) agents in Scenario I under Non-Reactive simulation setting under NMPC (top), CMPC (middle) and DRCMPC (bottom) solutions. We observe collision under NMPC, while the latter two do not result in collision.}}
    \label{fig:scenario_1_visualization}
\end{figure*}

%\subsection{Results} 

\subsubsection{Scenario I (Single Obstacle)}
We initialize the ego agent at position \((x_{\text{ego}}^{\text{init}}, y_{\text{ego}}^{\text{init}}) = (0, 0.5)\) and the obstacle agent at \((x_{\text{obs}}^{\text{init}}, y_{\text{obs}}^{\text{init}}) = (10, 0.5)\). The target velocity for the ego agent is set to \(v_{\text{ego}}^{\text{target}} = 15\) m/s, and the obstacle agent's target destination remains \((x_{\text{obs}}^{\text{final}}, y_{\text{obs}}^{\text{final}}) = (400, 0.5)\). The yaw angle for both the ego and obstacle agents is initialized to \(\psi_{\text{ego}} = 0\), \(\psi_{\text{obs}} = 0\), and the initial velocity for the obstacle agent is \(v_{\text{obs}} = 1\) m/s. We further set $x_{\max} = [\infty, 2, 2\pi, 15]$, $x_{\min} = [-\infty, 0, 0, 0]$.

The trajectories of the ego and obstacle agents under NMPC, CMPC and DRCMPC are shown in the top, middle and bottom rows of Figure \ref{fig:scenario_1_visualization}. Both agents are moving from left to right on the same lane. Under NMPC, the ego agent performs lateral motion to avoid a collision; however is not successful in doing so since as the ego velocity reduces, the velocity of the reactive obstacle reduces even further following \eqref{eq:scenario1_velocity} which is not captured under the constant velocity assumption. Under CMPC, the ego agent is barely able to avoid collision. DRCMPC, on the other hand, executes the most conservative maneuver by maintaining a sufficiently large safety margin by initiating lane change much before CMPC and NMPC solutions.

\begin{figure*}[ht]
\centering
    \includegraphics[width=0.9\textwidth]{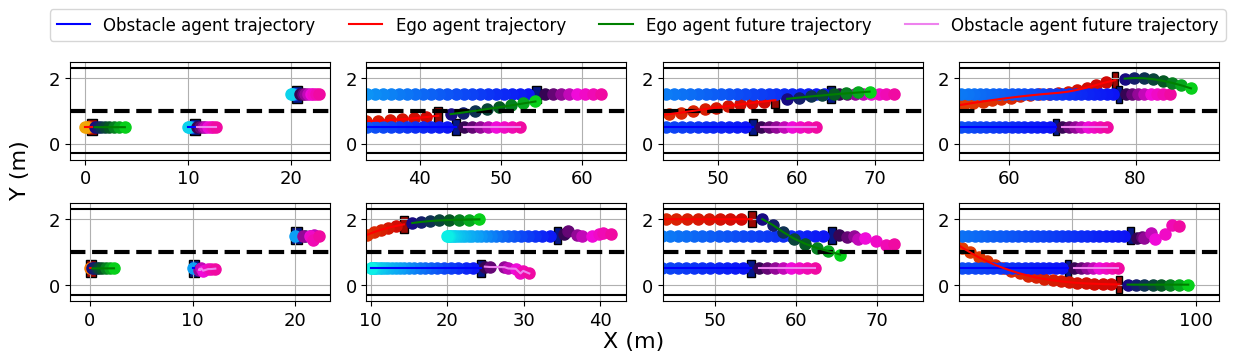}
    \caption{\footnotesize{Actual and predicted trajectories of the ego (red) and obstacle (blue) agents in Scenario II under Non-Reactive simulation setting under NMPC (top) and DRCMPC (bottom) solutions. We observe collision under NMPC, while DRCMPC does not result in collision.}}
    \label{fig:scenario_2_visualization}
 \end{figure*}

\subsubsection{Scenario II (Multiple Obstacles)}
We initialize the ego agent at position \((x_{\text{ego}}^{\text{init}}, y_{\text{ego}}^{\text{init}}) = (0, 0.5)\) and two obstacles at \((x_{\text{obs}_1}^{\text{init}}, y_{\text{obs}_1}^{\text{init}}) = (10, 0.5)\) and \((x_{\text{obs}_2}^{\text{init}}, y_{\text{obs}_2}^{\text{init}}) = (20, 1.5)\) respectively. The target velocity for the ego agent is set to \(v_{\text{ego}}^{\text{target}} = 15\) m/s, and the obstacle agent's target destination remains \((x_{\text{obs}_1}^{\text{final}}, y_{\text{obs}_1}^{\text{final}}) = (400, 0.5)\) and \((x_{\text{obs}_2}^{\text{final}}, y_{\text{obs}_2}^{\text{final}}) = (400, 1.5)\) with a maximum achievable velocity of 10 m/s. The yaw for the ego and obstacle are initialized to \(\psi_{\text{ego}} = 0\), \(\psi_{\text{obs}_1} = 0\) and \(\psi_{\text{obs}_2} = 0\), with the obstacle agent's initial velocity \(v_{\text{obs}_1} = 1\) m/s and \(v_{\text{obs}_2} = 1\) m/s. We further set $x_{\max} = [\infty, 2, 2\pi, 15]$, $x_{\min} = [-\infty, 0, 0, 0]$. 

Figure \ref{fig:scenario_2_visualization} shows that NMPC fails to avoid collision while DRCMPC is able to execute collision-free motion for the ego agent. Note that the training data was collected from a reactive obstacle which aimed to avoid collision, while in the above simulations, the obstacle is non-reactive. Despite the change in behavior of the obstacle, the DRCMPC approach is successful in avoiding collision.

\subsection{Monte Carlo Simulations}
We now repeat the above two scenarios $25$ times each for each of the three approaches. In each instance, we randomly choose the initial state of the obstacle agent; specifically, we choose the initial $x$-position and velocity of the obstacle from a Gaussian distribution. We compare the performance of the proposed approaches against non-reactive/neutral, cooperative and adversarial obstacle agents. We start by examining the impact of the radius of the ambiguity set $\epsilon$ towards satisfying the safety specifications. 

\begin{figure}[ht]
\centering
    \includegraphics[width=0.42\textwidth]{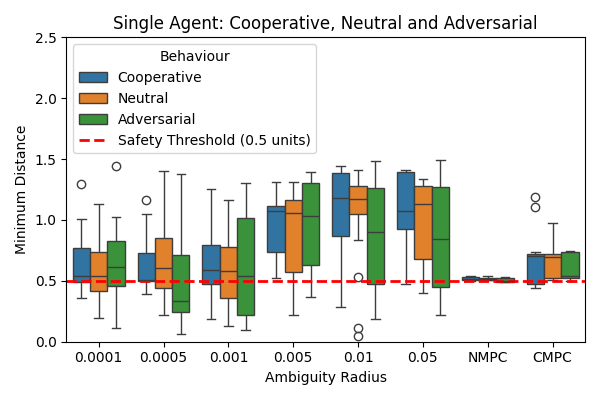}
    \caption{\footnotesize{Box plot of individual minimum distance between the ego and the single obstacle agent for different MPC techniques, obstacle behavior, and radius of the ambiguity set in DRCMPC ($\epsilon$).}}
    \label{fig:ambiguity_radius}
\end{figure}

\begin{figure}[ht]
\centering
    \includegraphics[width=0.42\textwidth]{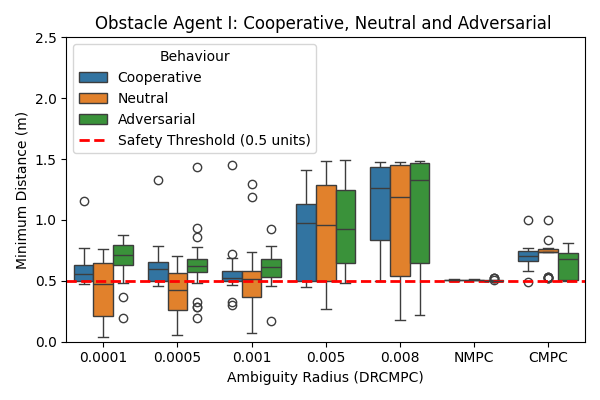}
    \caption{\footnotesize{Box plot of individual minimum distance between the ego and one of the two obstacle agents for different MPC techniques, obstacle behavior, and radius of the ambiguity set in DRCMPC ($\epsilon$).}}
    \label{fig:ambiguity_radius_multi}
\end{figure}

\subsubsection{Safety Guarantees} \label{monte_carlo_safety_gurantees}

Figure \ref{fig:ambiguity_radius} shows the box plot of the minimum distance between the ego agent and the obstacle agent for different types of obstacle behavior and the solution approach adopted by the ego agent for the general cost function stated in \eqref{eq:mpc_cost}. The results obtained under DRCMPC for a large range of radius of the ambiguity set ($\epsilon$) are compared with the results obtained under NMPC and CMPC techniques. The figure shows that as $\epsilon$ increases, the minimum distance increases as well; this is expected since a larger $\epsilon$ requires the solution to be robust against a larger set of possible distributions leading to a more conservative or safe trajectory for the ego. Under the NMPC approach, the minimum distance hovers around the safety threshold, while under the CMPC approach, it slightly larger than the threshold. When the obstacle behaves in a reactive manner by including the collision avoidance constraint in its MPC formulation, the minimum distance tends to be larger compared to when the obstacle is non-reactive and does not try to avoid collision with the ego agent. 

Our previous observations are consistent with the multi agent simulation setting as shown in Fig. \ref{fig:ambiguity_radius_multi} for one of the two agents. The results for the second obstacle are nearly identical and are omitted in the interest of space. Overall, our simulations indicate that while NMPC exhibits lateral movement, it often fails to ensure strict collision avoidance. CMPC is more successful in avoiding collision, but often executes suboptimal maneuver. DRCMPC, with a suitable choice of $\epsilon$, consistently performs smooth lateral movement while maintaining the safety distance effectively.

To validate the presence of distributional shifts across different obstacle behavior, we compute the empirical MMD distance (with RBF kernel) between the trajectories generated under these behaviors. These results shown in Table \ref{tab:divergence} confirm that the divergence between the distribution of trajectories under similar behavior is significantly smaller compared to the divergence between trajectories under different behavior. The empirical divergence is mostly comparable to the values of $\epsilon$ chosen in our simulations.

\begin{table}[h!]
\centering
\caption{Empirical MMD distance between different obstacle behavior}
\label{tab:divergence}
\begin{tabular}{|c|c|}
\hline 
\textbf{Scenario} & MMD Distance\\
\hline
Cooperative vs. Cooperative & 0.00054 \\
Neutral vs. Neutral     & 0.00029 \\
Adversarial vs. Adversarial & 0.00029 \\
Cooperative vs. Neutral & 0.6175 \\
Cooperative vs. Adversarial & 1.6708\\ 
\hline
\end{tabular}
\end{table}

\begin{figure}[ht]
\centering
    \includegraphics[width=0.42\textwidth]{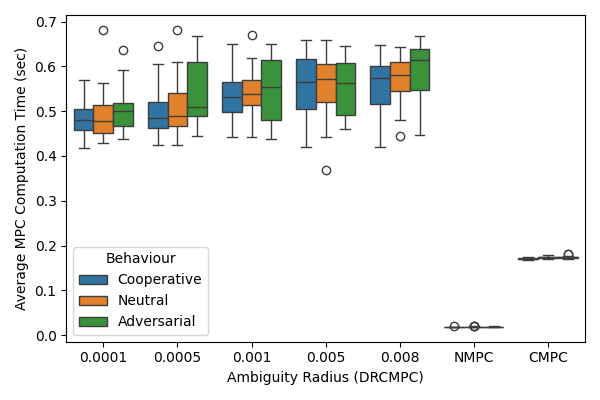}
    \caption{\footnotesize{Box plot of computation time for each instance of MPC for different MPC techniques, obstacle behavior, and radius of the ambiguity set in DRCMPC ($\epsilon$).}}
    \label{fig:time}
\end{figure}

\subsubsection{Computation Time} \label{monte_carlo_time_complexity}
Figure \ref{fig:time} shows the box plot of computation time for solving each MPC instance under each of the three methods considered in this work and for different obstacle behavior. As expected, the computation time taken by DRCMPC is much larger compared to the computation time taken by the NMPC and CMPC techniques. A larger $\epsilon$ leads to a larger computation time since the constraints become more restrictive. The distributionally robust conditional value at risk constraint \eqref{eq:dro_cvar_safety} is the most significant contributor to the computation time. Indeed, the reformulation of \eqref{eq:dro_cvar_safety} shown in \eqref{after_approx} shows that the first constraint is a non-convex conic constraint, and the second constraint scales linearly as the number of data points which together result in a comparatively larger computation time. We conducted additional numerical simulations (not reported in this letter due to space constraints) which showed that computation time scales approximately linearly with respect to $N_s$. Furthermore, when we used a linear (double integrator) dynamics instead of the bicycle model for the plant, the computation time reduced, but the reduction was not very significant.

Note that our Python-based high-level implementation of MPC is not optimized for real-time implementation, and a native C++ implementation would considerably reduce the computation time for each of the three techniques compared to the time reported here. If the computation time is still not acceptable, one could adopt a hierarchal approach where at the higher level, the MPC problem is solved for a large enough sampling time which allows computation to converge, and at the lower level, a control-barrier function (CBF) based safety filter \cite{hsu2023safety} is deployed to prevent collision in between two sampling times of the MPC. The CBF based safety filter requires a low dimensional quadratic program to be solved, and hence can be deployed at a much smaller sampling time ($\sim 10 \quad \mu$s). Finally, note that data-driven distributionally robust optimization problems are in general computationally expensive. Nevertheless, the rigorous guarantees provided by such techniques have inspired many researchers to investigate them in this safety-critical application \cite{navsalkar2023data,hakobyan2019risk,akella2024risk}. The increase in computation time is the price to pay for enhanced safety guarantees.
%%%%%%%%%%%%%%%%%%%%%%%%%%%%%%%%%%%%%%%%%%%%%%%%%%%%%%%%%%%%%%%%%%%%%%%%%%%%%%%%

\section{Conclusion}
\label{conclusion}

We presented a novel data-driven collision avoidance approach which incorporates contextual information via the conditional kernel mean embedding operator and distributional robustness via an ambiguity set defined in terms of the MMD distance. Simulation results show that the proposed approach is able to compute a collision-free trajectory for the ego agent in several challenging scenarios and different possible obstacle behavior. There remain several promising directions for future research, such as developing computationally tractable reformulations and performing more extensive training and simulations that capture urban traffic scenarios.

\bibliographystyle{IEEEtran}
\bibliography{main_arxiv}

\end{document}